\documentclass{article}

\usepackage{microtype}
\usepackage{graphicx}
\usepackage{subcaption}
\usepackage{booktabs} % for professional tables
\usepackage{float}

\usepackage{hyperref}

\newcommand{\mycomment}[1]{}

\usepackage[accepted]{icml2026}

\usepackage{amsmath}
\usepackage{amssymb}
\usepackage{mathtools}
\usepackage{amsthm}

\usepackage[capitalize,noabbrev]{cleveref}

\theoremstyle{plain}

\theoremstyle{definition}

\newtheorem*{oproblem*}{Open Problem}

\theoremstyle{remark}

\usepackage{tcolorbox}
\usepackage{footnote}
\BeforeBeginEnvironment{tcolorbox}{\savenotes}
\AfterEndEnvironment{tcolorbox}{\spewnotes}
\icmltitlerunning{Position: Building a Science of AI}

\begin{document}
	
	\twocolumn[
	\icmltitle{Position: Don't Just ``Fix it in Post'':\\ A Science of AI Must Study Training Dynamics}
	
	
	
	\icmlsetsymbol{equal}{*}
	
	% Author order is currently arbitrary and subject to change based on people's bandwidth
	%
	% Other people contacted: Will
	
	\begin{icmlauthorlist}
		\icmlauthor{Stella Biderman}{eai}
		\icmlauthor{Mohammad Aflah Khan}{mpi}
		\icmlauthor{Niloofar Mireshghallah}{cmu}
		\icmlauthor{Catherine Arnett}{eai}
        \icmlauthor{Fazl Barez}{ox}
        \icmlauthor{Naomi Saphra}{bu,hv}

	\end{icmlauthorlist}
	
	\icmlaffiliation{eai}{EleutherAI}
	\icmlaffiliation{cmu}{Carnegie Mellon University}
	\icmlaffiliation{bu}{Boston University}
    \icmlaffiliation{hv}{Harvard University}
	\icmlaffiliation{ox}{University of Oxford, Martian}
	\icmlaffiliation{mpi}{Max Planck Institute for Software Systems}
	\icmlcorrespondingauthor{Stella Biderman}{stella@eleuther.ai}
	
	\icmlkeywords{Deep Learning, Science of AI, Scaling Laws, Interpretability, Learning Dynamics, Philosophy of Science}
	
	\vskip 0.3in
	]
	
	\printAffiliationsAndNotice{}  % no special notice (required even if empty)

\begin{abstract}

    What would it mean to have a \textit{scientific} understanding of AI? Models are not static objects: they are snapshots of time-evolving processes shaped by data, objectives, architectures, and optimization dynamics. Yet much of AI research treats models as fixed artifacts, analyzing behaviors after training rather than asking why they emerge. This position paper argues that a science of AI must move beyond post-hoc fixes and study the training dynamics that produce model behavior. Such a science should support progressively stronger forms of understanding: predicting outcomes from early training signals, intervening when trajectories go wrong, and ultimately designing training procedures that more reliably produce desired properties. Scaling laws have made prediction routine for loss; the challenge is extending this success to capabilities, biases, robustness, and safety-relevant behaviors. We articulate requirements for such theories grounded in the history and philosophy of science, examine progress in mechanistic interpretability, fairness, memorization, and simplicity bias, and identify concrete open problems.
\end{abstract}

\section{Introduction}\label{sec:intro}

    Consider a tennis ball thrown into the air. Knowing the ball's position, velocity, and acceleration is sufficient to predict its motion with stunning accuracy. Add in higher order derivatives and a relatively simple differential equation will give the position of the ball until the end of the universe, with the primary source of error being the level of precision with which we can record the state of the rest of the stuff in the universe at the moment that the ball is thrown. This predictive power was hard-won, grounded in developing an understanding of how forces like gravity and properties like inertia operate on objects. The equations of motion are not merely curves fit to past observations, they are consequences of a deeper theory about how the universe works on a fundamental level that gives rise to the physical phenomenon we see in the world around us every day.

	What would it mean to have such a theory for AI? %It would be a causal account of how training dynamics produce model behaviors. It would provide a unified framework for how all aspects of model training influence model development. Such a theory would not only describe the training process, but identify the underlying mechanisms. %And that understanding would enable us to predict, intervene in, and ultimately design training processes---not merely analyze their outputs after the fact.
    \textbf{In this position paper, we argue that the field is far from having a scientific theory of AI and must move beyond post-hoc fixes to study the training dynamics of models in order to develop one.} 
    When models exhibit undesirable behaviors, the dominant response is to ``fix it in post''---to apply post-training interventions like RLHF or input and output filtering---rather than seek to understand why the behavior emerged. By studying models as static artifacts, we miss the opportunity to come up with explanations about how aspects of training affect the final model. Understand how training works as a dynamic process will allow us to predict and design effective training procedures.

\subsection{The ``Fix It in Post'' Mentality}

	In film production, the phrase ``fix it in post'' describes a common cost-cutting strategy: deferring problems during shooting and correcting them in post-production. In AI, an analogous mentality has taken hold. Large models are trained on massive datasets with limited curation and, when they don't behave as desired, people ask ``how can I further finetune the model to eliminate this undesirable behavior?'' rather than ``why did the model develop this behavior and how could it have been avoided?''

	The limitations of this approach are almost immediately apparent. There is extensive work detailing how safety finetuning can be shown to be highly brittle, such as by finetuning on benign data \citep{qi2023fine,lo2024largelanguagemodelsrelearn}, asking questions in the past tense \cite{andriushchenko2024does}, or asking them in a low resource language \citep{yong2023low}. Not only is the state-of-the-art training approach for any given task to do finetuning, but it is vanishingly rare that any other effective methodologies exist. Despite the extensive effort to combat frequency biases in models, the underlying representational biases persist \citep{mickel2026more}. This highlights the fundamental limitations of approaching AI behavior as something to be patched rather than something to be understood.

    \begin{tcolorbox}[colback=blue!3,colframe=blue!40,sharp corners,title={\textbf{Example: Band-Aids for Racism}}]
		When DALL-E 2 generated mostly white men for prompts like ``CEO,'' OpenAI's fix was to  secretly appending words like ``Black'' or ``female'' to user prompts \citep{baio2022evidence,offert2022sign}. This produces superficially diverse outputs while leaving the underlying bias intact. This is highly reminiscent of issues with Google Photos, which still cannot recognize gorillas over a decade after Black users were infamously misclassified as apes \citep{grant_google_photo_gorillas_2023}. After years of unsuccessful remediation, Google ultimately disabled primate recognition altogether rather than figure out how to fix the root cause \citep{guardian_google_gorilla_2018}.
	\end{tcolorbox}

	This pattern parallels the history of other scientific disciplines, where  intervention has preceded understanding and failures were met with more aggressive tuning rather than deeper inquiry. Pre-germ-theory medicine optimized treatments without causal models of disease \citep{de1926microbe}; behaviorist psychology shaped behavior while refusing to theorize about internal representations \citep{behaviourchomsky1959chomsky}; early aviation compensated for instability with brute force rather than aerodynamics \citep{aerovincenti1990engineers}. In each case, empirical success masked a lack of explanatory mechanism. 
    
    History suggests that such phases do not  end simply through refinement of the interventions themselves, but when latent mechanisms are articulated and the space of possible behaviors becomes constrained by theory. Early flight experiments largely relied on trial-and-error, with designers making incremental adjustments to wing shapes and control surfaces without a principled understanding of lift or stability \citep{aerovincenti1990engineers}. The Wright brothers advanced the field by systematically measuring aerodynamic forces in wind tunnel experiments, producing detailed empirical tables of lift and drag \citep{wright1901some}. However, a full theoretical account lagged behind. Only with the development of modern aerodynamics, based on fluid dynamics and informed by Isaac Newton’s laws of motion, did flight become a predictable engineering discipline rather than an experimental art \citep{prandtl1923applications, anderson2011ebook}. What changed was not simply optimizing to the desired outcome, but investing in a fundamental understanding of the underlying mechanisms that constrained the design space.
    
    We argue that the current dominant paradigm in machine learning research seeks primarily to optimize models to perform well at specific benchmarks without contributing to a deeper understanding model training, and that the dominance of this attitude is a major blocker to developing scientific theories of AI.

\section{What would a ``Science of AI'' Look Like?}

    One natural approach to envisioning a science of AI is to look to what is considered good practice in fields of science and see what can be imported into our field. In \textit{the Essential Tension} \citep{kuhn1977essential}, the philosopher of science Thomas Kuhn laid out six characteristics of good scientific practice.\footnote{In the original Kuhn identifies five characteristics, combining external and internal consistency. We follow subsequent work and distinguish between them.}

    \paragraph{Empirical Accuracy} A scientific theory should be able to make correct empirical predictions about the world.

    \paragraph{Internal Consistency} A scientific theory's claims and assumptions should not contradict one another.

    \paragraph{External Consistency} Scientific theories should be consistent with other existing  scientific theories.

    \paragraph{Scope}  A scientific theory's consequences should extend beyond the particular observations, laws, or phenomena it was originally designed to explain.

    \paragraph{Simplicity} A scientific theory should rely on as few independent assumptions as possible. Note that this is \textit{conceptual simplicity}, in the sense of \citet{aristotle_posterior} and \citet{aquinas1912summa}, among others and not syntactic simplicity.

    \paragraph{Fruitfulness} A scientific theory should disclose new phenomena or previously unrecognized relationships among known ones, opening productive lines of research beyond those it was built to address.

    Immediately we see significant divergence from typical research in AI, so much so that it doesn't make sense to try to judge work by AI researchers along these dimensions. Most research in AI doesn't produce theories that are intended to provide predictive accuracy across diverse experimental contexts or reveal underlying truths about the way the world works. To build a science of AI, then, we must develop our practices so that it makes sense to judge AI research by the standards that one judges fields of science. In the rest of this section we describe conceptual approaches to doing AI research that we believe are key to setting the field on a more scientifically grounded course.

%\subsection{Requirements for a Science of AI}\label{sec:requirements}

%What does it look like to design a scientific paradigm for AI? We believe that there are three core heuristics that AI researchers can focus on to improve their scientific practices: predictive theories (Section \ref{sec:prediction}), studying training dynamics (Section \ref{sec:dynamics}), and generalization across models (Section \ref{sec:generalization}). %In addition to encouraging scientific thinking on their own, these three heuristics also empower the \textit{predict}, \textit{intervene}, \textit{design} pipeline we view as a path forward for bringing about a science of AI. \naomi{at this point in the paper, I don't really understand what the connections are between these two different ontologies. I think we need to either make the connection up here, or we need to reconsider trying to relate these two different things.} 

\subsection{Prediction, Not Just Description}
\label{sec:prediction}

    Arguably the most important innovation in the history of science was the shift to center the ability to make predictions about \textit{what will happen} should an experiment be run, and then providing the framework for verifying the prediction. Sometimes this takes the form of precise equations (e.g., the laws of motion in physics) and sometimes this takes the form of less exact heuristics and methodologies that nevertheless can be used to draw reliable inferences (e.g., the theory of evolution by natural selection in biology). These predictions have significant practical utility, but they're also the primary way by which we can judge the empirical accuracy of a theory or demonstrate it has a broad scope. When a theory makes correct empirical predictions about data we haven't seen yet, we can have stronger confidence in the theory's empirical accuracy. When a theory tells us what will happen in a context we haven't previously considered, we have an opportunity to judge how broad the scope is and (potentially) check its external consistency with the predictions of theories that were developed in that context. In the study of AI, we rarely make such predictions. We describe, in great detail, how a model behaves \textit{after} training. But \textit{before} training, we rarely ever make claims about what we expect a model to do.

    \begin{tcolorbox}[colback=blue!3,colframe=blue!40,sharp corners,title={\textbf{Example: Darwin's Moth}}]
		In 1862, Darwin encountered the orchid \textit{Angraecum sesquipedale}, with a nectar spur nearly 30cm long. He reasoned that if the orchid reproduces something must pollinate it, and that only an insect with a proboscis long enough to reach the nectar would do. He predicted such a moth must exist in Madagascar. Over four decades later, the moth was discovered, confirming the prediction \citep{arditti_2012_moth}.
	\end{tcolorbox}

	One notable exception is predicting pretraining loss with scaling laws \citep{hestness2017deep,rosenfeld2019constructivepredictiongeneralizationerror,kaplan2020scaling,hoffmann2022training}. Given model size and a compute budget, final loss can be predicted with reasonable accuracy. Labs use this to plan training runs and to flag issues early in a training run if it isn't in line with predictions, which offers practical utility.\footnote{We are not aware of any clear public writings on this topic, and this comment is based on personal communication.} Recent work has also shown success at extending prediction to some evaluation benchmarks \citep{held2026relativescalinglawsllms,chen2024scaling,grattafiori2024llama,achiam2023gpt}. We can conceptually extend this paradigm to more generally ask, for a particular property $P$, what methodologies would enable us to predict the value of $P$ for a model given smaller models. While some research in this direction exists for memorization \citep{biderman2023emergent,prashanth2024recite} and social bias \citep{biderman2023pythia,patel2025fairness}, we view the ability to develop these theories as an essential component of the science of AI.

\subsection{Models as Time-Evolving Processes}
\label{sec:dynamics}

    Scientific theories provide answers to ``why'' questions. While fields such as interpretability often claim to answer such questions for AI, typical work misses something crucial. Answering the question ``why did a model do X on Y input'' certainly has some utility (e.g. for corporations interested in product assurance, user-satisfaction, and compliance), but on a scientific level it's fundamentally limited. A more scientific mindset would be to ask ``why did the model develop this behavior?'' This involves shifting from viewing models as static objects to viewing them as snapshots of time-evolving processes and studying the entire dynamical system \cite{saphra2023interp, biderman2023pythia,hoogland2023towards}. When the object of study is the training process rather than the finished model, an account of that process can be applied any model it produces, not to one specific set of weights \citep{sellam2021multiberts}.

    One approach examines training step-by-step. \citet{olsson2022context} discovered that induction heads form abruptly during a phase transition early in training. \citet{saphra2019understanding} showed that representations develop in a characteristic order, with syntax emerging before semantics and \citet{chen2023sudden} showed that internal syntactic structure precipitates the acquisition of grammatical rules. \citet{kangaslahti2025hidden} found that smooth aggregate loss curves hide discrete skill-specific breakthroughs, visible only when loss is decomposed by sample type.
    
    A complementary approach varies training conditions across runs. \citet{qin2024sometimes} train models on data with varying syntactic complexity, finding that data composition determines whether models learn hierarchical rules or surface shortcuts. A number of results have likewise found that models can only systematically compose concepts which appear in diverse contexts during training \citep{allen2023physics,okawa2023compositional,chang2025characterizing}.
    
    Understanding these training trajectories should matter, even to people solely interested in the final model performance, because it enables \textit{intervention during training}. If a model is trending toward memorizing sensitive content, developers could adjust the data before memorization solidifies. If an undesirable bias is emerging, the data mixture could be re-balanced \citep{biderman2023pythia}. \citet{biderman2023emergent} showed that intermediate checkpoints predict final memorization better than smaller fully-trained models, suggesting that the trajectory contains information that endpoint comparisons miss.

\subsection{Properly Identifying Objects of Study}
\label{sec:generalization}

    Every scientific theory must choose what to treat as a phenomenon to be explained, what to consider a confounder, and what to treat as primitive. Is temperature a parameter to a thermodynamic system, or is it a measure of particle movement? Do we inherit eye color, or do we inherit genes which control the proteins that determine eye color? As science advances, the primitives of one theory become the results of a deeper one.

    In AI we face analogous choices, but these choices are often made implicitly rather than deliberately \citep{ayonrinde2025mathematical}. When we study a phenomenon across models with different random seeds, we are implicitly declaring that seed-specific details are not part of our explanation: they are noise to be averaged over, not signal to be accounted for. %When we study models across scales, we treat scale as a variable while holding architecture fixed. These choices define the scope of our theories: what they can explain and what they take as given.

    %We point this out not because any of these choices are necessarily bad or made incorrectly, but because thinking systematically about them is essential for developing theories that are simple and fruitful. In all likelihood there are better and worse choices one could make in various contexts, but by failing to attend to the importance of this we prevent ourselves from performing the analysis required to tell.

    %\catherine{I think there needs to be a better link between these two paragraphs} \stella{Tigges et al. isn't a good citation here... One of Naomi's papers would be better.}
     %The result of a single training run cannot distinguish between regularities of the learning process and accidents of initialization, and only provides information about a specific collection of weights. %A finding reported from a single model with a single random seed has not yet committed to any level of abstraction. Our claims do not apply to our proposed method or setting---only to a single instantiation, a specific collection of weights. Such findings cannot distinguish between regularities of the learning process and accidents of initialization. 
     This is not merely a theoretical concern. Even high-level findings can be sensitive to random seeds \citep{sellam2021multiberts}, and behaviors identified in one model may not appear in another trained on slightly different data \citep{qin2024sometimes}. Near ``emergence'' thresholds, random variation leads to clusters of generalization behavior \citep{zhao2024distributional} which each correspond to different loss basins \citep{juneja2022linear} and internal mechanisms \citep{li2025can}. Sparse autoencoders trained with different random initializations can learn different features \citep{paulo2025sparse}, revealing limitations for using them as a proxy for a full model. Without systematic studies of variation, we cannot know whether we are describing the results of fundamental phenomena or implementation details.

        \begin{tcolorbox}[colback=blue!3,colframe=blue!40,sharp corners,title={\textbf{Example: Data Attribution}}]
		Given a trained model, its training data, and a behavior of interest, the goal of data attribution is to identify the training examples most responsible for that behavior. Data encountered later in training has a larger influence on model behavior, so the answer depends on what is held fixed: a method that conceptualizes the task as reshuffling the data and retraining \citep{ilyas2022datamodels,park2023trak} and one that holds the realized run fixed and asks what that specific model owes to each example \citep{ilyas2025magic} are answering different questions. Unfortunately, its common for work to confuse or conflate these settings \citep{deng2024texttt,mlodozeniec2025distributional,wang2026better}.  
        \end{tcolorbox}

    This logic extends beyond random seeds to other dimensions of variation. Most NLP research studies English-centric models, yet claims to study ``language models'' rather than ``English language models''.  This common oversight led to the ``Bender Rule'' \citep{bender2019rule}: that research in natural language processing should explicitly state which languages are studied. 
    
    For models where training data are released or described, training data often consists of at least 90\% English\footnote{Increasingly, powerful models are trained on high proportions of English and Chinese data, e.g. DeepSeek \citep{guo2025deepseek}, but the same issue remains as claims are rarely limited to English-Chinese language models.} data \citep{brown2020gpt3,grattafiori2024llama}, but claims about them are usually about learning language generally. At the same time, research on non-English languages is seen to be uninformative to field more broadly.
    %
    %If findings about syntax or semantics do not generalize across typologically diverse languages, they may reflect properties of English rather than properties of learning. 
    In fact, recent results suggest that models trained in exactly the same way learn some languages more easily than others \citep{cotterell-etal-2018-languages, arnett-bergen-2025-language}, and model design decisions impact different languages differently \citep{gerz-etal-2018-relation,arnett2025explaining,shani2026roots}. 
    A deliberate science of AI would make explicit choices about which dimensions of variation matter for which questions, and then systematically test those choices.

    \subsection{Predict, Intervene, Design: Progress towards our Goals for a Science of AI}

    How can we tell if we are making progress towards developing a science of AI? We cannot measure a theory the way we can measure a benchmark score, so we need some concrete way to tell whether we are acquiring one.
    
    A useful move is to ask what such an understanding should empower us do, and to treat our growing ability to do it as a measure of progress. We identify three capabilities, each more demanding than the last and each building on the one before. The first is \textit{predicting training}: forecasting a property of a model from its training setup, before the run finishes and ideally before it begins, requiring that we can make reliable predictions (Section 2.1). The second is \textit{intervening on training}: recognizing that a trajectory is heading somewhere undesirable and redirecting it while training is still underway, requiring an understanding of models that views them as a time-evolving dynamical process (Section 2.2). The third is \textit{designing training}: fixing the properties we want in advance and constructing a procedure that reliably produces them, requiring that we have correctly identified which behaviors are stable enough to target (Section 2.3). Each is harder than the last because each requires everything the one before it did, and then more.

    These capabilities operationalize progress, but its important to remember that they do not define the goal. The goal is a scientific theory that explains why training produces the behaviors it does. The capabilities themselves are not the goal but instead are how we would recognize that we are making progress towards a predictive theory. We could predict a property by fitting a curve to past runs without understanding what drives it, and we could in principle design a procedure that yields a desired property by searching over enough configurations, with no account of why it works. Abilities like these are necessary signs of understanding, since a theory that predicts nothing and guides no intervention is hardly a theory. They are not sufficient for it. A predictor or a recipe that explains nothing is precisely the kind of opaque success this paper argues against.

    Treating the capabilities as progress rather than as the destination also clarifies why the program is worth pursuing before any complete theory arrives. Each is valuable on its own as we acquire it: a reliable prediction saves a failed run, a timely intervention keeps a harmful behavior from setting in, and even partial design makes outcomes more controllable. These are the returns collected along the way, and they are why a science of training dynamics is not a luxury to be deferred until the theory is finished.

    The case studies in \Cref{sec:case-studies} can be read as a progress report of this kind, measuring different areas against the three capabilities. Mechanistic interpretability includes some of the most promising work here, studying how circuits form across training and confirming their role through intervention. But its mainstream is heavily influenced by the production needs of companies and by anxieties about advanced AI toward compliance-style accounts of why a finished model produced a given output. Fairness is further behind and seems largely uninterested in catching up. Memorization is another field that has seen a lot of success building towards a scientific approach lately, with work focused on making and validating predictions and on developing theory-driven understanding of why memorization happens. Work on simplicity bias is assembling the mechanistic pieces an explanation would need. Measured this way, progress is real but uneven, concentrated in a few areas and barely begun in others.

	\section{Case Studies}
	\label{sec:case-studies}

	We now examine several research areas through our proposed lens, identifying promising directions and significant gaps in contemporary research practice.

	\subsection{Mechanistic Interpretability}

	Mechanistic interpretability aims to reverse-engineer neural network computations, identifying circuits responsible for specific behaviors. This goal aligns naturally with scientific aspirations: identifying mechanisms is precisely the causal, explanatory work that distinguishes science from description.
	However, most mechanistic interpretability work remains fundamentally descriptive. It answers ``what computations does this model perform?'' rather than ``why did these computations develop?'' or ``how can we instill specific mechanisms?''

	The first step towards answering these questions is to find generalizable patterns. Most circuit analysis studies a single model---often a single checkpoint. When circuits are identified in GPT-2 Small, we typically do not know whether they exist in GPT-2 Medium, in a different random seed, or at intermediate training steps.
	\citet{tigges2024llm} directly address this gap, studying circuit consistency across checkpoints and scales. Their finding is nuanced: high-level mechanisms are consistent, but which specific neurons participate fluctuates considerably. Similarly, \citet{riviere2025start} find that the layers of the attention heads most responsible for sense disambiguation performance varies across random seeds. If individual neuron participation is unstable, then ``neuron 347 in layer 8 detects indirect objects'' is the wrong level of description---it captures an implementation detail rather than a stable phenomenon. A scientific theory must characterize mechanisms at abstraction levels that are invariant. And in a variety of settings, even the high-level mechanisms can vary widely, leading to variation in their resulting edge case behavior \citep{li2025can,juneja2022linear,huang2025measuring}.

    By studying mechanisms across multiple settings (per Section \ref{sec:generalization}), researchers can confirm hypotheses about their connection to specific model behavior. \citet{chen2023sudden}, noting that language models developed internal syntactic structure suddenly, manipulated the timing of this onset to confirm that it precipitated a subsequent breakthrough in grammatical capabilities. \citet{li2025can} trained hundreds of small models, correlating specific attention patterns with intuitive out-of-distribution behaviors. All of these approaches confirmed a link between the mechanism and its associated behavior.

    One success story for training analysis (as proposed in Section \ref{sec:dynamics}) is arguably the most well-studied mechanism in LLMs: induction heads. \citet{olsson2022context} first showed that induction heads formed alongside an increase of in-context learning capabilities. Subsequent results have deepened and complicated this link further; in-context learning can be limited to an early stage of training \citep{singh2023transient} and as LLMs memorize each task in their weights, they rely less on induction heads to execute it \citep{yin2025attentionheadsmatterincontext}. When an association is thus confirmed throughout training and across different settings, we can be confident in the link.

    But most work in mechanistic interpretability neglects the ultimate cause of observed structures. Without such confirmation, interpretability claims can be misleading. For example, selective neurons (i.e. neurons that activate for specific output classes; \citealp{zhou2014object}) were a promising early candidate for interpretability through monosemanticity. Unfortunately, these monosemantic neurons were revealed to damage model performance \citep{leavitt2020selectivityconsideredharmfulevaluating}, a result explained when \citet{ranadive2023specialroleclassselectiveneurons} showed them to be a vestigial remnant of early training. Rather than providing a viable interpretability method in performant models, selectivity revealed a side effect of the training process.  
    
    %\todo{talk about interp illusions \citep{bolukbasi2021interpretability} and OOD stuff here?}

    % Interpretability illusions are especially concerning when explanations are used to predict out-of-domain behavior. A plausible explanation need not be a faithful account of the computation that will persist under distribution shift. \citet{bolukbasi2021interpretability} show that BERT neurons and linear directions can appear to encode simple concepts, even when the apparent concept is partly an artifact of embedding-space geometry and the narrow support of the analyzed corpus. Similarly, \citet{adebayo2018sanity} demonstrate that saliency maps can remain visually plausible after model parameters or labels are randomized, suggesting that some explanations track input structure or method priors rather than learned computation. \citet{dombrowski2019explanations} further show that explanations can be manipulated by imperceptible input perturbations while preserving the model's prediction, indicating that explanation stability is not guaranteed even locally.

    \begin{tcolorbox}[colback=violet!3,colframe=violet!40,sharp corners,title={\textbf{Open Problem: Differentiating Between Computation and Data Structure}}]
    One simple way of using the training process in interpretability is to compare a trained model with its random initialization. Both SAEs \citep{heap2025sparse} and saliency maps \citep{adebayo2018sanity} provide plausible interpretations at initialization, calling these methods into question as ways of understanding a trained model. These results also reveal a general problem when analyzing structure in hidden representations: which structures are surface-level patterns in the input data, which actually illuminate the model's processing of those inputs, and which are best ascribed to minds of the people doing the research \citep{bolukbasi2021interpretability,meloux2025dead,ayonrinde2025mathematical}? 
	\end{tcolorbox}

    Interpretability methods can also create an illusion of understanding when explanations are only validated on the same distribution that produced them. \citet{bolukbasi2021interpretability} show that individual neurons and linear directions in BERT can appear to encode simple semantic concepts, even when those apparent concepts reflect the geometry of the representation space and the narrow support of the datasets being inspected. In this sense, an interpretation can be locally compelling while failing to identify a stable mechanism. A related problem appears for feature-attribution explanations: \citet{dombrowski2019explanations} show that explanations can be arbitrarily manipulated by small input perturbations that leave the model's output nearly unchanged, tying this instability to the geometry of neural-network decision surfaces. Both results suggest that explanations should not be trusted merely because they are sparse, semantically appealing, or visually plausible.

    Out-of-domain generalization provides a natural stress test for such claims. If an interpretation identifies a real mechanism rather than a dataset artifact, it should predict behavior not only on the examples used to discover it, but also under controlled distribution shifts where the hypothesized mechanism is preserved and competing shortcuts are broken. This point connects mechanistic interpretability to the broader literature on shortcut learning: models often achieve high in-domain accuracy by relying on features that are predictive in the benchmark but non-causal for the task \citep{Geirhos_2020}. \citet{mccoy-etal-2019-right} show that NLI models trained on standard datasets can rely on syntactic heuristics that fail during controlled evaluation, despite strong in-domain performance. Similar concerns apply to explanations themselves: \citet{chrysostomou-aletras-2022-empirical} find that common faithfulness metrics for post-hoc explanations can behave misleadingly in out-of-domain settings, motivating explicit random baselines and distributional tests. Thus, interpretability claims should be evaluated as scientific hypotheses: they should generate predictions about when a model will generalize, when it will fail, and which interventions would change that behavior.

    % \catherine{We need some examples to show when people are not doing the right thing. Ideally a famous case where a prediction made based on a single model/checkpoint/seed failed to generalize. Right now, at the end of the section, you feel like people are already generally doing the right thing.}\stella{An Interp. Illusion for BERT}

    % \aflah{Taken together, these results suggest that mechanistic interpretability has not yet converged on the level of abstraction required for scientific explanation. While some high-level mechanisms generalize across scale, neuron-level implementations do not, and worse, many interpretability methods recover convincing explanations even in randomly initialized networks \citep{méloux2025deadsalmonsaiinterpretability}. A scientific account of neural mechanisms must therefore demonstrate robustness to scale, initialization, and analytical method which are criterion that existing work do not yet meet.}

	\subsection{Fairness and Bias}
    
    An extensive literature documents how biases in training data propagate into learned models. Word embeddings exhibit gender stereotypes in occupational associations \citep{bolukbasi2016man}; facial recognition systems show dramatically different error rates across demographic groups \citep{buolamwini2018gender}; recidivism algorithms have higher false positive rates for minoritized racial groups \citep{angwin2016machine}. This work has been valuable for awareness and evaluation metrics, yet it remains largely diagnostic. Practitioners measure bias after training and attempt mitigation through resampling, reweighting, or constrained optimization, but cannot answer the forward-prediction question of \Cref{sec:prediction}: given a target distribution, what training data modifications would produce a model exhibiting it?
	
    Without predictive theories mapping data characteristics to model behavior, fairness interventions remain empirical trial-and-error. Practitioners cannot reason forward from dataset design to downstream consequences, heavily limiting the potential impact of this work. \citet{sellam2021multiberts} and \citet{biderman2023pythia} demonstrate how their model suites can enable more rigorous scientific study of these phenomena, but the gender bias literature is yet to build on them.%This demonstrates that controlled interventions on training data \textit{can} affect downstream bias, but the precise mapping from corpus statistics to bias metrics remains unclear. The authors even note that the reliability of existing bias measures is an open question. Such work illustrates the \textit{kind} of controlled experimentation a science of bias requires, while highlighting how far we remain from predictive theory. This is the mid-training intervention of \Cref{sec:dynamics}: changing the data as bias emerges rather than patching the finished model.

    \begin{tcolorbox}[colback=violet!3,colframe=violet!40,sharp corners,title={\textbf{Open Problem: Designing Social Bias}}]
		Suppose one decides \textit{a priori} that an image generation model should generate a woman when prompted to generate a person of a given profession with frequency equal to the U.S. Bureau of Labor Statistics' measured data. What should the rate of women in the training data be to accomplish this? Setting it equal to the target rate is insufficient \citep{zhao2017men,seshadri2024bias,chen2024would,roos2025met}, but there is currently no training method that results in the desired behavior.
	\end{tcolorbox}
    
    We know models exhibit biases, and we know training data contains imbalances, but we lack studies that quantitatively connect the two across the full pipeline. \citet{hall2022systematic} show that bias amplification occurs primarily when recognizing group membership is easier than recognizing class membership, and that amplification correlates with model capacity and overconfidence---but this work focuses on discriminative classifiers in controlled settings. For LLMs, each training stage could amplify, preserve, or dampen input bias through different mechanisms: pretraining may bake in statistical associations \citep{bender2021dangers}, while RLHF can either mitigate bias through preference learning or introduce new biases from annotator demographics \citep{kirk2023understanding}. Recent work suggests that bias introduced during pretraining persists through fine-tuning \citep{ghate2025intrinsic}, indicating that \textit{post hoc} interventions may be fundamentally limited. Furthermore, as synthetic data becomes a larger fraction of training corpora \citep{alemohammad2023self}, we need to understand whether synthetic generation launders bias (making it harder to detect) or compounds it through feedback loops \citep{mehrabi2024flirt}. Without causal tracing through the pipeline, fairness interventions remain guesswork.

\begin{tcolorbox}[colback=violet!3,colframe=violet!40,sharp corners,title={\textbf{Open Problem: Cultural and Linguistic Erasure}}]
	Model behavior varies significantly across languages, dialects, cultures, and even names associated with different backgrounds, resulting in uneven product quality and downstream harms \citep{blodgett2020language}.
    Generative models can flatten or distort marginalized cultures, and hallucinations can create misinformation about communities, histories, or practices \citep{ortu2026preservinghistoricaltruthdetecting}. Yet we lack systematic measurement infrastructure: which cultures are most prone to erasure or distortion? 
    Can we predict, before deployment, which cultures and communities will experience representation harms, and how these harms vary with model scale, training data composition, and post-training choices?
\end{tcolorbox}

% The Global MMLU benchmark reveals that 28\% of questions require culturally sensitive knowledge, and models optimized for translated benchmarks risk overfitting to Western-centric concepts \citep{singh2025global}. Addressing this requires both better evaluation infrastructure---benchmarks that test cultural knowledge across many communities---and causal studies connecting training choices to downstream cultural appropriateness.

Global MMLU is a step toward this kind of measurement infrastructure: it finds that 28\% of questions require culturally sensitive knowledge, and shows that models optimized for translated benchmarks risk overfitting to Western-centric concepts \citep{singh2025global}. But evaluation alone does not explain where these failures enter the pipeline. Addressing cultural and linguistic erasure therefore requires both better benchmarks that test cultural knowledge across many communities and causal studies connecting training choices to downstream cultural appropriateness.

% \catherine{I think there could be a nice paragraph here to highlight how languages and cultures do not have a one-to-one mapping. Sometimes we want crosslingual transfer (factual knowledge) and sometimes we don't (culture-specific knowledge, e.g. what is the emergency number). Sometimes we do, e.g. performance on a task in the target language that relies on knowledge of another culture even if all the training data that discuss that knowledge is in a different language.}
% Multilingual capability does not imply multicultural alignment: improved language performance does not correlate with better representation of that language's associated cultural values \citep{rystrom2025multilingual}. 
% \begin{itemize}
%     \item In some cases, language-specific information will not transfer across languages \citep{goldman2025eclektic}
%     \item Sometimes, there is asymmetric transfer between langauges \citep{zhang-etal-2025-cross}
%     \item Questions in the same language may differ between cultural contexts (Who is the president? What is the emergency number?). 
% \end{itemize}

\subsection{Memorization}

    Memorization research offers a compelling example of nascent theory-building, with recent work establishing both predictive frameworks and causal methodologies that exemplify the scientific approach we advocate. The memorization case study in Pythia \citep{biderman2023pythia} explicitly framed their work as scientific prediction: they hypothesized, based on intuitions about training dynamics, that data encountered later in training would be more likely memorized. Initial experiments suggested order-independence, but subsequent work with refined methodology confirmed the original hypothesis \citep{lesci2024causal,kuditipudi2025blackbox}. This progression exemplifies the scientific process we advocate: (1) hypothesis stated \textit{before} experiments, (2) grounded in a causal story about training dynamics, (3) initial null result led to refined methods rather than abandonment, (4) finding generalizes across models. This is an instantiation of treating models as time-evolving processes (\Cref{sec:dynamics}): a claim about the training trajectory is stated up front and tested across models and through different versions.
    
    A second line of work has established rigorous methodology for studying \textit{how much} repetition is required for verbatim memorization. \citet{huang2024demystifying} developed a causal intervention framework to study verbatim memorization in controlled settings. Using cross-model interchange interventions, they show that non-trivial repetition is necessary for true memorization and that it cannot be attributed to specific weights alone, but is fundamentally tied to general language modeling capabilities. Apparent cases of ``single-shot'' memorization are often reconstructions of templated texts \citep{prashanth2024recite} or frequent patterns rather than genuine memorization. Additionally, over half of the interventions producing memorized tokens rely only on general language modeling components, explaining why unlearning attempts often degrade overall model quality. These interchange interventions transfer across models, exemplifying the generalization recommendations of \Cref{sec:generalization}.

    Membership inference studies provide converging evidence. \citet{duan2024membership} found that attacks across Pythia models barely outperform random guessing, due to massive datasets, near-one-epoch training, and fuzzy member/non-member boundaries. This suggests that, at scale, most sequences are not memorized in ways that leave detectable traces, consistent with findings that substantial repetition is required for verbatim memorization which is further supported by findings from the Hubble models \citep{wei2025hubble}.
    
    Despite the significant effort put towards developing causal analyses of memorization, significant and very basic gaps remain in our understanding of memorization dynamics. For example:
    
    % \begin{oproblem*}[Predicting Memorization]
    %     Given a dataset in a fixed order and suite of models of different sizes trained on said dataset, can you predict \textit{which specific sequences} will be memorized by a large model trained on that data?
    % \end{oproblem*}

    % This problem is introduced by \citet{biderman2023emergent}\footnote{Prior to this paper, the memorization literature overwhelmingly viewed memorization as a corpus-level statistical phenomenon rather than analyzing individual samples.}, who develop a baseline methodology for making the prediction and analyze their predictions by extrapolating memorization behavior from lower-compute trial runs to forecast which sequences a larger model will memorize. Despite only proposing a baseline methodology that they openly admit is highly inadequate, other than \citet{liu2023llm360} and \citet{dentan2024predicting} who replicate their results, we are unaware of any progress on this problem.

\begin{tcolorbox}[colback=violet!3,colframe=violet!40,sharp corners,title={\textbf{Open Problem: Predicting Memorization}}]
	\citet{biderman2023emergent} introduce a new question about memorization: given a dataset in a fixed order, can you predict \textit{which specific sequences} will be memorized by a large model trained on that data without training the full model? \citet{biderman2023emergent} analyze using both smaller and partially trained models to forecast memorization, concluding that their method was insufficient and challenge future work to do better. While this work has been replicated \citep{liu2023llm360} and influenced subsequent work on related questions \citep{dentan2024predicting,lesci2024causal,prashanth2024recite}, the core question is still open.
\end{tcolorbox}

% This problem is proposed by \citet{biderman2023emergent}, who develop a baseline methodology using scaling laws: extrapolating memorization behavior from lower-compute trial runs to forecast which sequences a larger model will memorize. They find that while recall (correctly predicting memorized sequences) follows consistent scaling patterns, precision remains anomalous---larger models exhibit behavior that deviates sharply from smaller models, particularly in low-compute regimes most useful for practical prediction. Subsequent work has identified correlates of memorization, including $n$-gram overlap and duplication rates \citep{carlini2023quantifying}, rare token presence \citep{prashanth2024recite}, and minimum repetition thresholds \citep{huang2024demystifying,borkar2025privacy,borkar2026memorization} as potential driving factors. Yet these factors explain variance \textit{post hoc}; we still lack predictive theories that can identify \textit{a priori} which specific sequences will be memorized given a training run specification.

% \begin{tcolorbox}[colback=violet!3,colframe=violet!40,sharp corners,title={\textbf{Open Problem: The Memorization--Capacity--Competence Triad}}]
% 		Can we predict memorization rates from the joint interaction of model capacity, linguistic competence, and data distribution dynamics throughout training?
% \end{tcolorbox}

Memorization is not simply a function of repetition and recency. Recent work reveals a surprising non-monotonicity: models are most ``absorbent'' around 10--20\% into training, despite having more unused capacity earlier \citep{duan2024membership}. Related evidence appears in \citet{masud-etal-2024-probing}, who find that finetuning early RoBERTa checkpoints works best for hate speech classification. This suggests an interplay between three factors: (1) remaining capacity, (2) linguistic competence (early models may lack the representational structure to encode verbatim sequences), and (3) data distribution (what else competes for the same capacity at each training stage). Additionally, \citet{liu2023llm360} show that memorization scores spike at the end of training and that earlier data is progressively forgotten unless revisited, while later checkpoints memorize more readily even for out-of-distribution sequences \citep{duan2024membership}. A unified theory predicting memorization rates from these factors jointly would enable principled decisions about when to introduce sensitive data during training, and whether capacity-limited models can be ``safely'' trained on data that larger models would memorize.
%These open problems focus on prediction---the first level of our hierarchy. Equally important are the subsequent levels: \textit{intervention} (detecting memorization mid-training and adjusting data or objectives before it solidifies) and \textit{design} (engineering data pipelines that prevent problematic memorization while preserving beneficial factual grounding from the start).

\subsection{Simplicity Bias and Learning Dynamics}

	Research on distributional simplicity bias represents a promising foundation for scientific theory. \citet{refinetti2023neural} established that networks trained with SGD learn statistical features of increasing complexity: first differences in class means, then covariance structure, only later higher-order cumulants like co-skewness. \citet{belrose2024neural} extended this to real data, demonstrating characteristic U-shaped loss curves confirming progressive sensitivity to higher-order statistics. More recently, \citet{chang2025bigram} provided mechanistic evidence that language models learn a bigram subnetwork early in training. This is consistent with findings in \citet{michaelov2025language}, which show that language model predictions correlate with unigram probabilities very early in training. Shortly after, they correlate with bigrams, then trigrams, and so on. These patterns hold across model scales and architectures.
    
	A related line studies spectral bias: the tendency to learn low-frequency functions before high-frequency ones \citep{rahaman2019spectral,xu2019frequency,basri2019convergence}. Both concern the order of feature learning, but are not straightforward generalizations of each other. Spectral bias concerns Fourier decomposition of the learned function; distributional simplicity bias concerns statistical moments of data distributions. It is natural to ask whether there is a potential unification of these theories. For example, it is possible that the real bias is simply towards patterns that occur in training data disproportionately often and \textit{nature} (or our data collection methods) have a bias towards patterns easily explained in terms of lower order statistics and low-frequency signals.

    \begin{tcolorbox}[colback=violet!3,colframe=violet!40,sharp corners,title={\textbf{Open Problem: simplicity bias $\to$ OpenFold}}]
	   The OpenFold paper \citep{ahdritz2024openfold} trains models to predict protein stuctures and finds something remarkable: the model seems to learn spatial dimensions sequentially over the course of training and the predictions produced by checkpoints saved early in training are approximately PCA optimal projections of the final prediction into lower dimensions. This seems quite evocative of research on simplicity biases, but its unclear which simplicity bias literature it connects to. Are the empirical results reported in \citet{ahdritz2024openfold} consistent with theoretical literature, and if so could these results have been predicted in advance?
	\end{tcolorbox}

    \section{Alternative Views}

    % \todo{add citations}

    Our argument that AI research should prioritize predictive, causal theories of training dynamics runs against several influential perspectives in the field. In this section, we outline these alternative viewpoints and explain where we agree, where we disagree, and what is ultimately at stake.

    \paragraph{Engineering Sufficiency.} A common view holds that AI does not require deeper scientific theory: large-scale experimentation, benchmarking, and post-hoc fixes are sufficient as long as systems improve empirically. This perspective is supported by the undeniable success of current engineering practice \citep{alexnet, resnet, brown2020gpt3, chowdhery2022palmscalinglanguagemodeling}. Our claim is not that such methods fail, but that they optimize locally and opaquely. Without predictive theories, failures cannot be anticipated, interventions do not reliably generalize, and success at one scale offers limited guidance at the next \citep{xu2025surveyattackslargelanguage, unreliable_scaling_law}. Engineering has made real but fragile advances, and by supplementing it with scientific rigor we can cement the gains from engineering-focused work in a more reliable and robust fashion.

    \paragraph{Anti-Theory Skepticism.} Some argue that neural networks are too complex for meaningful theory, or that the stochastic nature of training makes it impossible \citep{baldassi2016unreasonable, pontin2018greedy, adolfi2024complexity}. We agree that AI theories might be approximate, heuristic, and domain-specific. Still, empirical regularities such as scaling laws, memorization dynamics, and simplicity bias show that even partial theories can yield predictive insight \citep{kaplan2020scaling, chen2023sudden, biderman2023emergent, biderman2023pythia, chen2024scaling, tao_2024_vocab_scaling_laws, held2026relativescalinglawsllms}. The key question is whether such theories constrain expectations and guide interventions better than post-hoc description, even if they remain imperfect. Furthermore, other fields such as thermodynamics, quantum mechanics, and genetics have shown that even highly non-deterministic processes can be fruitfully analyzed through statistical tools. %\catherine{Relatedly, I feel like reviewers are so not used to reviewing theoretical work that they usually penalize papers for not being empirical enough, even if it's stated that the paper's contribution is theoretical.}

    \paragraph{Automation as Scientific Progress.} A growing view holds that the path to AI understanding is to automate research itself: using AI systems to generate hypotheses, run experiments, and discover patterns that may be opaque to humans \citep{lu2024aiscientist, aiscientist_v2}. On this account, scientific insight need not be human-interpretable as long as automated processes reliably produce better models. While such tools can accelerate discovery, this stance risks conflating optimization with understanding. If the research process itself becomes a black box, we gain neither predictive theories nor principled control. Instead, there is faster iteration on artifacts whose failure modes remain poorly understood.

    \paragraph{Safety Pragmatism.} In safety and alignment work, an influential view prioritizes immediate harm reduction through post-training interventions \citep{ouyang2022traininglanguagemodelsfollow, bai2022constitutionalaiharmlessnessai}, arguing that waiting for deeper understanding is impractical given deployment pressures. This concern is legitimate, and short-term mitigations are often necessary. However, many recurring safety failures (e.g. jailbreaks, regressions, brittleness under scale) are a direct result of the ``fix it in post'' attitude \citep{wei2023jailbrokendoesllmsafety, zou2023universaltransferableadversarialattacks} and offer no path toward design-level solutions that would prevent problems from arising. A science of training dynamics would not replace pragmatic safeguards but supplement them with more robust methods as they are developed, such as \citet{obrien2025deep} and \citet{cloud2024gradient}.

	\section{Conclusion}

	We have argued that AI research should aspire to genuine scientific theories: causal explanations that predict novel phenomena, generalize across instances, and unify disparate observations. The field's current emphasis on \textit{post-hoc} analysis and post-training fixes---the ``fix it in post'' mentality---while sometimes practically necessary, cannot substitute for fundamental understanding.

	The building blocks exist. Scaling laws demonstrate that prediction is achievable: we can forecast loss from early signals and design compute allocations accordingly. Work on simplicity bias shows causal mechanisms can be identified. Released checkpoints and datasets enable the longitudinal studies that dynamical understanding requires.
	What is needed is extending scaling laws' success to the properties we care about. We envision a theory that allows us to first \textit{predict} outcomes from early training signals, then \textit{intervene} when trajectories go wrong, ultimately \textit{design} training procedures that guarantee desired properties. Scaling laws have achieved this for loss; the challenge is achieving it for capabilities, biases, and safety.

	The problems we have outlined are difficult---but they are the problems that a mature science of AI will solve.

	\section*{Acknowledgments}

    This work was informed by conversations with Jennifer Mickel, Aviya Skowron, Isabelle Lee, and Louis Jaburi.

    N.S. is supported by a Technical AI Safety Research Grant from Coefficient Giving via Berkeley Existential Risk Initiative. This work has been made possible in part by a gift from the Chan Zuckerberg Initiative Foundation to establish the Kempner Institute for the Study of Natural and Artificial Intelligence at Harvard University.
	\bibliography{main}
	\bibliographystyle{icml2026}
	
\end{document}